\documentclass[runningheads]{llncs}

\usepackage{graphicx}
\usepackage{amsmath}
\usepackage{amssymb}
\usepackage{mathtools}
\usepackage{booktabs}
\usepackage{xcolor}
\usepackage{multirow}

\usepackage{hyperref}[colorlinks=true]
\hypersetup{hidelinks, backref=true, pagebackref=true, hyperindex=true, breaklinks=true, colorlinks=true, urlcolor=blue, bookmarks=true, bookmarksopen=false, pdftitle={Title}, pdfauthor={Author}}

\newcommand{\N}{\mathcal{N}}
\newcommand{\U}{\mathcal{U}}
\newcommand{\G}{\mathcal{G}}
\newcommand{\V}{\mathcal{V}}
\newcommand{\F}{\mathcal{F}}
\newcommand{\E}{\mathcal{E}}
\newcommand{\R}{\mathbb{R}}
\newcommand{\tab}{\hspace{0.5cm}}

\newcommand{\abs}[1]{|#1|}
\newcommand{\norm}[1]{\Vert #1 \Vert}

\newcommand{\gray}[1]{\textcolor{black}{#1}}
\newcommand{\SO}[1]{\mathrm{SO}(#1)}
\newcommand{\ball}[1]{\mathcal{B}_r(#1)}

\renewcommand{\subsection}[1]{\subsubsection{#1}}

\begin{document}

\title{Mesh Convolutional Neural Networks for Wall Shear Stress Estimation in 3D Artery Models}
\titlerunning{Mesh convolutional networks for wall shear stress estimation}
\author{Julian Suk\inst{1} \and Pim de Haan\inst{3,2} \and Phillip Lippe\inst{2} \and Christoph Brune\inst{1} \and \\Jelmer M. Wolterink\inst{1}}  %
\authorrunning{Suk et al.}
\institute{Department of Applied Analysis \& Technical Medical Centre, University of Twente, Enschede, The Netherlands \\ \email{j.m.suk@utwente.nl} \and QUVA Lab, University of Amsterdam, Amsterdam, The Netherlands \and Qualcomm AI Research, Qualcomm Technologies Netherlands B.V.\thanks{Qualcomm AI Research is an initiative of Qualcomm Technologies, Inc.} \\  %
}
\maketitle

\begin{abstract}
	Computational fluid dynamics (CFD) is a valuable tool for personalised, non-invasive evaluation of hemodynamics in arteries, but its complexity and time-consuming nature prohibit large-scale use in practice. Recently, the use of deep learning for rapid estimation of CFD parameters like wall shear stress (WSS) on surface meshes has been investigated. However, existing approaches typically depend on a hand-crafted re-parametrisation of the surface mesh to match convolutional neural network architectures. In this work, we propose to instead use mesh convolutional neural networks that directly operate on the same finite-element surface mesh as used in CFD. We train and evaluate our method on two datasets of synthetic coronary artery models with and without bifurcation, using a ground truth obtained from CFD simulation. We show that our flexible deep learning model can accurately predict 3D WSS vectors on this surface mesh.
	Our method processes new meshes in less than 5 [s], consistently achieves a normalised mean absolute error of $\leq 1.6$~[\%], and peaks at 90.5~[\%] median approximation accuracy over the held-out test set, comparing favourably to previously published work. This demonstrates the feasibility of CFD surrogate modelling using mesh convolutional neural networks for hemodynamic parameter estimation in artery models.
\end{abstract}

\section{Introduction}
In patients suffering from cardiovascular disease, parameters that quantify arterial blood flow could complement anatomical measurements such as vessel diameter and degree of stenosis. For example, the magnitude and direction of wall shear stress (WSS) was found to correlate with atherosclerotic plaque development and arterial remodeling~\cite{SamadyEshtehardi2011,HoogendoornKok2019}. %
Imaging techniques like particle image velocimetry or 4D flow MRI with which WSS could be measured non-invasively are not widely available and may be less accurate in smaller arteries. Therefore, WSS is often estimated using computational fluid dynamics (CFD). This requires the extraction of a geometric artery model from e.g. CT images, spatial discretisation of this model in a finite-element mesh, and iterative solution of the Navier-Stokes equations within the mesh~\cite{TaylorFonte2013}.
However, high-quality CFD solutions require fine meshes, leading to high computational complexity.
Therefore, there is a practical demand for surrogate models that trade accuracy for speed. %

\begin{figure}[t]
	\centering
	\includegraphics[width=\textwidth]{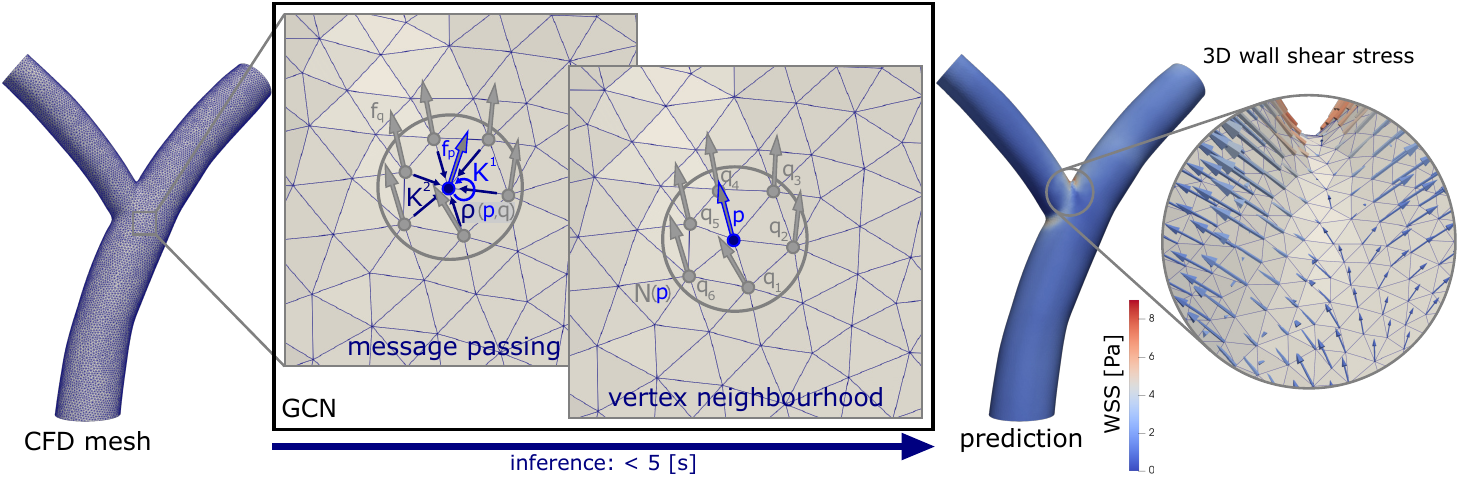}
	\caption{Our graph convolutional network (GCN) predicts vector-valued quantities on a computational fluid dynamics (CFD) surface mesh in a single forward pass. It consists of convolutional layers that apply filters with kernels $K^1, K^2$ in a vertex neighbourhood $q \in N(p)$ to vertex features $f_p, f_q$ using parallel transport $\rho(p, q)$ (``message passing").}
	\label{fig:overview}
\end{figure}

Deep neural networks are an attractive class of surrogate models. They require generating training data (e.g. by CFD) but once they are trained, inference in new geometries only requires a single forward pass through the network, which can be extremely fast.
Previous applications of deep learning to coronary CFD surrogate modelling have studied the use of multilayer perceptrons to estimate fractional flow reserve based on hand-crafted features~\cite{ItuRapaka2016}, as well as the use of convolutional neural networks to estimate WSS based on 1D parametrisations of single arteries~\cite{SuZhang2020} and 2D views of bifurcated arteries~\cite{GarleghiSamarasinghe2020}. Furthermore, convolutional neural networks have been applied in stress and hemodynamics prediction in the aorta~\cite{LiangLiu2018,LiangMao2020}. 

All of these approaches have in common that they rely on a global or local \textit{parametrisation} of the artery model, which is limited to idealised vessels or does not generalise to more complex vessel trees. This limits their value as a plug-in replacement for simulation in clinical CFD workflows.
In contrast, we propose to let deep neural networks learn a (latent) parametrisation of the vessel geometry directly from a surface mesh representing the vessel lumen wall.
We describe input meshes as graphs and use graph convolutional networks (GCNs) and their extension, mesh convolutional networks, to predict WSS vectors on the mesh vertices (Fig.~\ref{fig:overview}). This offers a plug-in replacement for CFD simulation operating on a mesh that can be acquired through well-established meshing procedures.

GCNs have been previously studied for general mesh-based simulations \cite{PfaffFortunato2021}. Moreover, several recent works have investigated the use of GCNs in the cardiovascular domain, e.g. for coronary artery segmentation \cite{WolterinkLeiner2019}, coronary tree labeling \cite{HampeWolterink2019}, or as a surrogate for cardiac electrophysiology models~\cite{MeisterPasserini2021}. In a very recent study, Morales et al.~\cite{MoralesFerezMill2021} showed that GCNs outperform parametrisation-based methods for the prediction of WSS-derived scalar potentials in the left atrial appendage. In this work, we compare isotropic and anisotropic mesh convolution for the prediction of 3D vector quantities on mesh vertices. Additionally, we show how the use of gauge-equivariant mesh convolution~\cite{HaanWeiler2021} results in neural networks that are invariant under translation and equivariant under rotation of the input mesh, two desirable properties for improved data efficiency. The latter is especially relevant in medical applications where access to large datasets is uncommon.
Our method is flexible and avoids the loss of information that may occur due to ad-hoc parametrisation of the artery model. We demonstrate the effectiveness of this approach on two sets of synthetic coronary artery geometries.

\section{Method}
We model the artery lumen wall as a 2D manifold in 3D Euclidean space. Its surface mesh is fully described by a tuple of vertices $\V$ and polygons $\F$, whose undirected edges $\E$ induce the graph $\G = (\V, \E)$.
We propose to use GCNs to predict WSS \textit{vectors} on the vertices of the graph. Those are defined as the tangential force exerted on the lumen wall by the blood flow. GCNs are trainable operators that map an input signal on a graph $f^\text{in}: \G \to \R^{c_\text{in}}$ to an output signal $f^\text{out}: \G \to \R^{c_\text{out}}$ on the same graph. Like many convolutional neural network architectures, they can be composed of convolutional and pooling layers, which are here defined as follows.

\subsection{Convolution}
We implement convolutional layers by spatial message passing in a neighbourhood $N(p)$ around vertex $p$, expressed in a general form as
\begin{equation}
	((K^1, K^2(\cdot, \cdot)) * f)_p \coloneqq f_p \cdot K^1 + \sum\limits_{q \in N(p)} \rho(p, q) f_q \cdot K^2(p, q), \tab p \in \V
	\label{eq:conv}
\end{equation}
where $K^1, K^2 \in \R^{c \times \bar{c}}$ are trainable kernel matrices. These matrices weigh the features of $p$ itself and those of its neighbours, respectively.
We let ${f_q = f(q) \in \R^c}$ denote the feature vector corresponding to $q \in \V$. Equation~(\ref{eq:conv}) describes three graph convolutional layers, depending on the choice of  $K^2(p, q)$ and ${\rho(p, q) \in R^{c \times c}}$ as follows.
First, we consider GraphSAGE~\cite{HamiltonYing2017} convolution with full neighborhood sampling, which corresponds to choosing an \textbf{isotropic} kernel $K^2(p, q) = \frac{1}{\abs{N(p)}} \bar{K}^2$ and $\rho(p, q) = I$, the $c \times c$ identity matrix. Isotropic kernels can
accommodate neighbourhoods $N(p)$ of a varying number of vertices without canonical ordering, but sacrifice expressiveness.

Second, we consider FeaSt~\cite{VermaBoyer2018} convolution, which implements an \textbf{anisotropic} filter through an attention mechanism $\rho(p, q) = \sigma(w \cdot (f_q - f_p)) I$ with trainable weights $w \in \R^c$ where $\sigma(\cdot)$ is the ``softmax" function. The kernel is again chosen $K^2(p, q) = \frac{1}{\abs{N(p)}} \bar{K}^2$.

Third, we employ gauge-equivariant mesh convolution~\cite{HaanWeiler2021}. Here, anisotropic kernels are learned as $K^2(p, q) = K^2(\theta^p_q)$, where the orientation angles $\theta^p \in [0, 2\pi)^{\abs{N(p)}}$ of all neighbours $q \in N(p)$ are measured from one \textit{arbitrary} reference neighbour projected onto the tangent plane at $p$.
This arbitrary choice defines a local coordinate system (``gauge"). Linearly combining vertex features $f_q$ requires expressing them in the same gauge,
which is done by the parallel transporter $\rho(p, q) = \rho(\gamma)$ where $\gamma(p, q)$ is the angle between the previously chosen reference neighbours. The representation $\rho(\cdot)$ is determined by the irreducible representations (``irreps") of the planar rotation group $\SO{2}$ which compose into the features $f_q, q \in \V$.
In order to ensure that the convolution does not depend on the arbitrary choice of local coordinate systems, a linear constraint is placed on the kernel $K^2$ such that the linearly independent solutions to this \textit{kernel constraint} span the space of possible trainable parameters.

\subsection{Pooling}
Graph pooling consists of clustering followed by a reduction operation.
We define clusters in the mesh based on a hierarchy of $n + 1$ vertex subsets $\V^0 \supset \V^1 \supset \dots \supset \V^n$
and assign each vertex in $\V^{i-1}$ to a vertex in $\V^{i}$ based on geodesic distances.
In the gauge-equivariant framework, we average features after parallel transporting to the same gauge. For unpooling, features are distributed over the clusters using parallel transport.
There is no notion of parallel transport for general $c$-dimensional features in GraphSAGE or FeaSt convolution, hence
we simply average over the $c$-dimensional feature space during pooling and distribute statically oriented features over the clusters during unpooling.

\subsection{Network architecture}
Since WSS depends on the global shape of the blood vessel, we use encoder-decoder networks, akin to the well-known U-Net. These networks are formed by three pooling scales where each scale consists of residual blocks with two convolutional layers and a skip connection. We copy and concatenate signals between corresponding layers in the contracting and expanding pathway. ReLU activation is used throughout the architecture.

\begin{figure}[t]
    \centering
    \includegraphics[width=\textwidth]{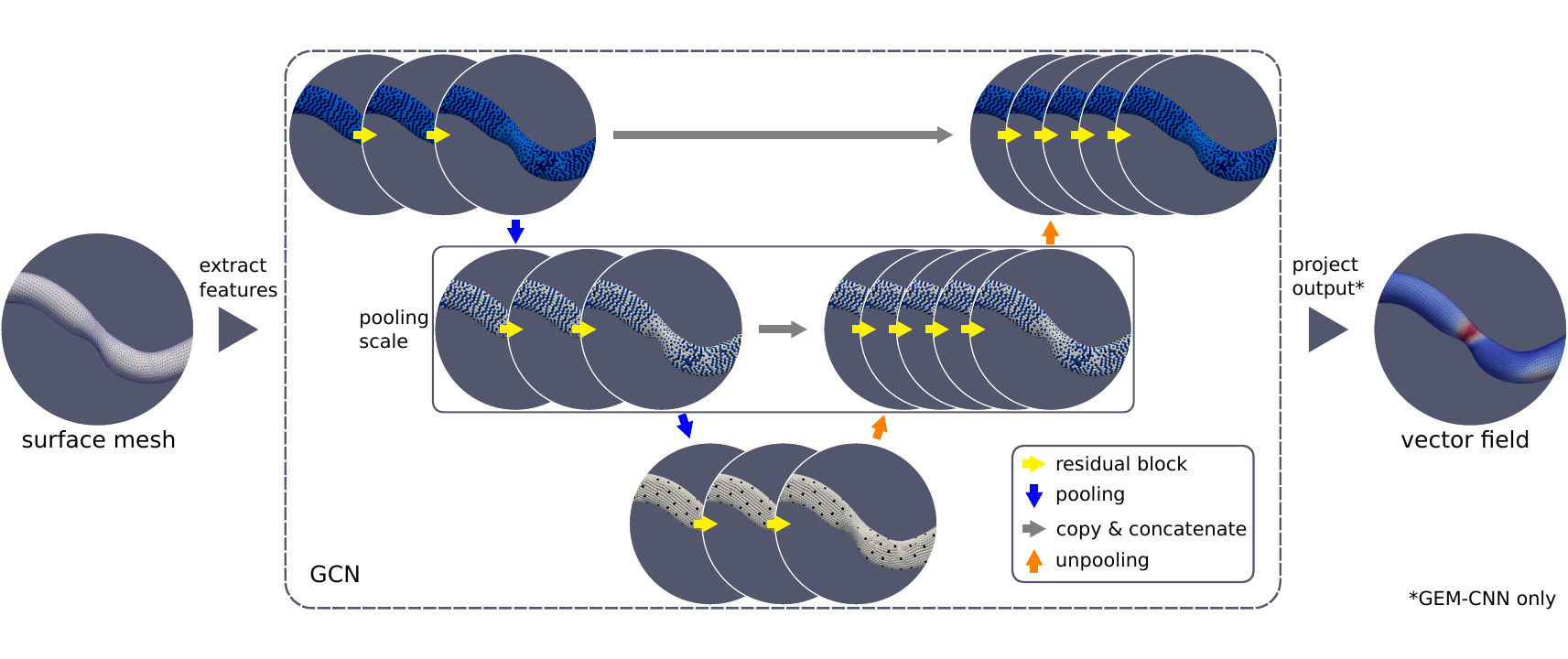}
    \caption{Schematic representation of the employed network architecture. Coloured vertices signify the ones used for message passing on each of the three pooling scales. Each residual block consists of two convolutional layers and one skip connection. For the gauge-equivariant network, the signal is projected back from the local coordinate systems into Euclidean space at the output.}
    \label{fig:architecture}
\end{figure}

We construct input features $f^\text{in}: \G \to \R^{c_\text{in}}$ that are invariant under translation and equivariant under rotation of the surface mesh.
This is done by, for each vertex $p$, considering a ball $\ball{p}$ and averaging weighted differences to the vertices $q \in \ball{p}$ as well as the vertex normals. Subsequently, we take the outer product of the averaged differences with themselves, the averaged normals with themselves, and the differences with the normals. For the GraphSAGE and FeaSt networks, we flatten and concatenate the resulting matrices and use the resulting feature vectors as inputs for each vertex. For the gauge-equivariant network, we instead decompose the matrices into factors of $\SO{2}$ irreps. A schematic representation of the employed architecture is shown in Fig.~\ref{fig:architecture}.

Gauge-equivariant convolutional layers are by construction equivariant under $\SO{2}$ transformation of the local coordinate systems.
Representing our network's input features $f^\text{in}$ in these gauges leads to end-to-end equivariance under $\SO{3}$ transformation of the global coordinate system. This means that the gauge-equivariant network is a globally rotation-equivariant operator.

Each network is additionally provided with a scalar input feature that contains, for each vertex, the (invariant) shortest geodesic distance to the vessel inlet, giving a rotation-equivariant indication of the flow direction.

\section{Datasets}
We synthesise two large datasets of coronary artery geometries and corresponding CFD simulations to train and validate our neural networks. The entire shape synthesis and CFD simulation is implemented and automated with the SimVascular~\cite{LanUpdegrove2018} Python shell.
\subsection{Single arteries}
We emulate the dataset by Su et al.~\cite{SuZhang2020} and synthesise 2000 idealised 3D coronary artery models. Each model has a single inlet and single outlet, with a centerline defined by control points in a fixed increment along the x-axis and a uniformly random increment along the y-axis. The cross-sectional lumen contour has a random radius $r \sim \U(1.25, 2.0)$ [mm]. Each artery contains up to two randomly-placed, asymmetric stenoses of up to 50 \%.
Each model is lofted and meshed using
a global edge length of 0.4 [mm], local edge length refinement proportional to the vessel radius, and five tetrahedral boundary layers. Fig.~\ref{fig:easy} shows an example synthetic geometry.
We simulate blood flow under the three-dimensional, incompressible Navier-Stokes equations. We assume dynamic viscosity  $\mu = 0.035$ [$\frac{\text{g}}{\text{cm} \cdot \text{s}}$], blood density $\rho = 1.05$ [$\frac{\text{g}}{\text{cm}^3}$], rigid walls, and no-slip boundary condition.
The inlet velocity profile is constant and uniform: $u_\text{in} = 20 [\frac{\text{cm}}{\text{s}}]$. A pressure $p_\text{out} = 100$ [mmHg] $\approx$ 13.332 [kPa] is weakly applied at the outlet.
A Reynolds number of $\text{Re} = 180 < 2100$ implies the fluid flow is laminar.
Blood flow is simulated until stationary and the simulation takes between 10 and 15 [min] on 16 CPU cores (Intel Xeon Gold 5218).

\subsection{Bifurcating arteries}
We synthesise 2000 3D left main coronary bifurcation models, randomly constructed based on an atlas of coronary measurement statistics~\cite{MedranoGraciaOrmiston2017}.
Each model consists of proximal main vessel (PMV), distal main vessel (DMV), and side branch (SB). Centerlines are defined by three angles sampled from the atlas. First, $\beta \sim \N(\mu_\beta = 78.9, \sigma_\beta = 23.1)$ [$^\circ$] is the angle between DMV and SB. Second,
$\beta' \sim \N(\mu_{\beta'} = 61.5, \sigma_{\beta'} = 21.5)$ [$^\circ$] is the angle between the bisecting line of the bifurcation and SB, describing skew of the bifurcation towards the child branch. Third,  $\gamma \sim \N(\mu_\gamma = 9.5, \sigma_\gamma = 21.5)$ [$^\circ$] is the angle under which the PMV centerline enters the bifurcation plane.
We model the vessel lumen with ellipses that are arbitrarily oriented in the plane normal to the centerline curve tangent. The lumen radii are drawn from the coronary atlas:
\begin{itemize}
    \item $r_\text{PMV} \sim \N(\mu_{r_\text{PMV}} = 1.75, \sigma_{r_\text{PMV}} = 0.4)$ [mm],
    \item $r_\text{DMV} \sim \N(\mu_{r_\text{DMV}} = 1.6, \sigma_{r_\text{DMV}} = 0.35)$ [mm], and
    \item $r_\text{SB} \sim \N(\mu_{r_\text{SB}} = 1.5, \sigma_{r_\text{SB}} = 0.35)$ [mm].
\end{itemize}
We sample lumen radii such that diameters follow a bifurcation law of the form $(d_\text{PMV})^a = (d_\text{DMV})^a + (d_\text{SB})^a + \delta$ with small $\delta$ and decrease approximately linear with vessel length.
Each model is lofted and meshed using
a global edge length of 0.2 [mm] and five tetrahedral boundary layers.
Blood flow simulation follows that of the single artery models, with differences as follows. We assume $\mu = 0.04$ [$\frac{\text{g}}{\text{cm} \cdot \text{s}}$], $\rho = 1.06$, and prescribe a parabolic influx profile corresponding to a uniform inflow velocity $u_\text{in} = 11.8 [\frac{\text{cm}}{\text{s}}]$ to get WSS values which agree with a healthy range~\cite{SamadyEshtehardi2011} of 1.0 to 2.5 [Pa]. We weakly apply a pressure $p_\text{out}$ at the union of the outflow surfaces. The Reynolds number for the fluid flow is $\text{Re} = 100 < 2100$ so the flow is laminar.
The simulation runs take between 8 and 12 [min] on 32 CPU cores (Intel Xeon Gold 5218).

\section{Experiments}
We implement three versions of the encoder-decoder network based on GraphSAGE, FeaSt and gauge-equivariant mesh convolution and refer to these as SAGE-CNN, FeaSt-CNN, and GEM-CNN. Hidden layers are set up so that each network has just under 800,000 trainable parameters.
Both datasets are split 80:10:10 into training, validation, and test set.
The GCNs are trained to predict a 3D WSS vector
at each vertex on the surface mesh. Since GEM-CNN naturally expresses its signal in the tangential planes w.r.t. each vertex, its output vector is here, for WSS, restricted to those tangential planes. However, outputting an additional normal component is also naturally supported. In our experiments, however, this only influenced the performance marginally. SAGE-CNN and FeaSt-CNN operate without the notion of tangential planes, thus they predict an unconstrained 3D vector to avoid overloading the methods and to save memory. The input meshes are the same surface meshes we use to obtain the ground truth via CFD, so that the CFD solution is directly used for network optimisation. All models are trained with a mean-squared error loss using the Adam optimiser.
SAGE-CNN and FeaSt-CNN are trained on an NVIDIA GeForce RTX 3080 (10 GB) for 400 epochs
taking 3:15 [h] and 12:30 [h] for the single artery models
and twice as long for the bifurcating artery models.
GEM-CNN is trained for 160 epochs (ca. 29:00 [h]) on two NVIDIA Quadro RTX 6000 (24 GB) for the single artery models, and 100 epochs (ca. 55:00 [h]) for the bifurcating artery models.
Inference in a new and unseen geometry takes less than 5 [s] including pre-processing. We open-source our PyTorch Geometric~\cite{FeyLenssen2019} implementation for the GraphSAGE and FeaSt networks.\footnote{\url{https://github.com/sukjulian/coronary-mesh-convolution}}

\begin{figure}[t]
    \centering
    \includegraphics[width=\textwidth]{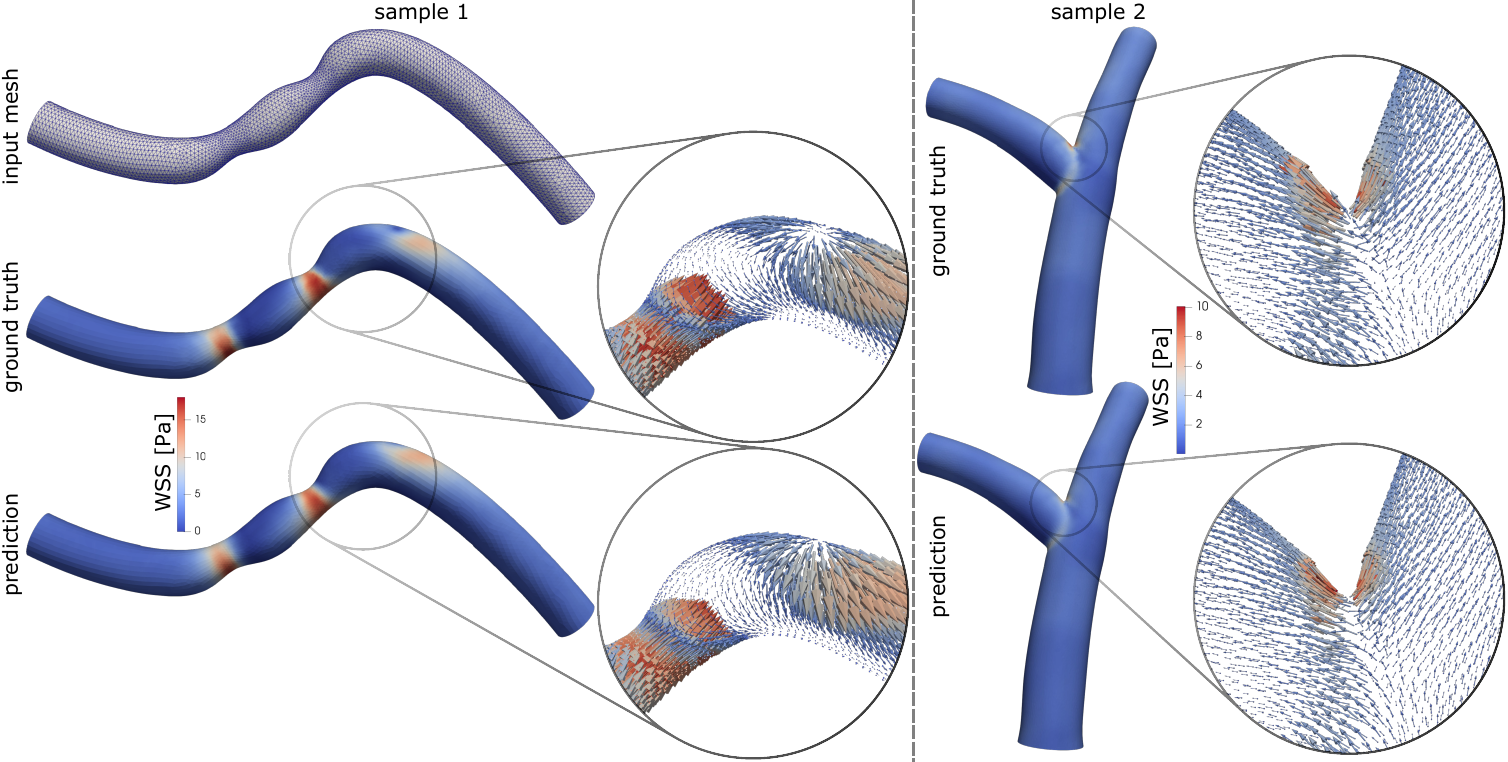}
    \caption{WSS prediction by our method (GEM-CNN) on two previously unseen artery samples. CFD simulation acts as ground truth. Our method predicts 3D WSS vectors for every mesh vertex. In sample 1, multidirectional WSS can be observed.}
    \label{fig:easy}
\end{figure}

We report mean absolute error normalised by the maximum WSS magnitude across the test set (NMAE) and define the approximation error $\varepsilon \coloneqq \norm{\triangle}_2 / \norm{L}_2$ where elements of $\triangle$ are vertex-wise normed differences between the network output $f^\text{out}: \G \to \R^3$ and ground truth $l: \G \to \R^3$ so that $\triangle_p = \norm{f^\text{out}(p) - l(p)}_2$ for $p \in \V$ and $L_p = \norm{l_p}_2$. Additionally, we report the maximum and mean vertex-wise normed difference, i.e. $\triangle_\text{max} = \max_p \{\triangle_p\}$ and $\triangle_\text{mean} = (\sum_p \triangle_p) / \abs{\V}$.

\subsection{Prediction accuracy}
Fig.~\ref{fig:easy} shows an example of the reference and predicted WSS in a single and a bifurcating artery model, illustrating that our method does not only predict the WSS magnitude but also the directional 3D vector. Table~\ref{tab:metrics} lists quantitative results of the prediction accuracy over both test sets.
Direct comparison of SAGE-CNN, FeaSt-CNN and GEM-CNN on the single artery models suggests strictly better performance when using anisotropic convolution kernels. GEM-CNN's good performance can be attributed to its anisotropic convolution kernels as well as the availability of parallel transport pooling. The results in Table~\ref{tab:metrics} are comparable to previously published results on a similar dataset~\cite{SuZhang2020}.
To demonstrate our method's flexibility, we train SAGE-CNN, FeaSt-CNN, and GEM-CNN on the bifurcating artery dataset without further adaption of the architecture. We find that all networks perform well, even without task-specific hyperparameter tuning. The performance differences between networks are less pronounced on the bifurcating arteries than on the single arteries. This hints at the input feature (parameters) in conjunction with the network's signal representation lacking expressiveness for the task at hand, since a convolutional neural network should always benefit from directional filters.

\begin{table}[tp]
	\begin{center}
		\resizebox{\columnwidth}{!}{
			\renewcommand{\arraystretch}{1.3}
			\begin{tabular}{@{}cccccccccccccccccc@{}}
				\toprule
				&&& \multicolumn{3}{c}{NMAE [\%]} & \phantom{abc} & \multicolumn{3}{c}{$\varepsilon$ [\%]} & \phantom{abc} & \multicolumn{3}{c}{$\triangle_\text{max}$ [Pa]} & \phantom{abc} & \multicolumn{3}{c}{$\triangle_\text{mean}$ [Pa]}\\
				\cmidrule{4-6} \cmidrule{8-10} \cmidrule{12-14} \cmidrule{16-18}
				&&& mean & median & $75$th && mean & median & $75$th && mean & median & $75$th && mean & median & $75$th\\ \midrule
				\multirow{6}{2cm}{\textbf{Single\\ Arteries}}
				& SAGE-CNN && 2.2 & 2.0 & 2.6 && 32.4 & 30.0 & 37.0 && 10.41 & 7.80 & 14.65 && 1.11 & 1.01 & 1.32\\
				& FeaSt-CNN && 1.2 & 1.1 & 1.5 && 19.0 & 18.6 & 22.4 && 5.83 & 5.13 & 8.17 && 0.60 & 0.57 & 0.77\\
				& GEM-CNN && \textbf{0.6} & \textbf{0.6} & \textbf{0.8} && \textbf{9.9} & \textbf{9.5} & \textbf{11.6} && \textbf{3.94} & \textbf{3.68} & \textbf{5.46} && \textbf{0.32} & \textbf{0.31} & \textbf{0.41}\\ \cmidrule{2-18}
				& SAGE-CNN$^\dagger$ && \gray{10.5} & \gray{9.6} & \gray{12.8} && \gray{149.2} & \gray{128.1} & \gray{181.2} && \gray{26.73} & \gray{23.96} & \gray{36.17} && \gray{5.31} & \gray{4.84} & \gray{6.50}\\
				& FeaSt-CNN$^\dagger$ && \gray{8.3} & \gray{7.5} & \gray{10.1} && \gray{123.7} & \gray{111.1} & \gray{152.9} && \gray{25.63} & \gray{22.93} & \gray{34.52} && \gray{4.22} & \gray{3.82} & \gray{5.13}\\
				& GEM-CNN$^\dagger$ && 0.6 & 0.6 & 0.8 && 9.8 & 9.4 & 11.4 && 3.80 & 3.39 & 5.53 && 0.32 & 0.31 & 0.42\\ \midrule
				\multirow{6}{2cm}{\textbf{Bifurcating\\ Arteries}}
				& SAGE-CNN && 1.3 & 1.1 & 1.5 && 24.4 & 21.1 & 27.3 && 4.38 & 4.14 & 5.50 && 0.27 & 0.22 & 0.29\\
				& FeaSt-CNN && \textbf{1.2} & \textbf{0.9} & \textbf{1.3} && \textbf{20.7} & \textbf{18.1} & \textbf{22.3} && 4.10 & 3.72 & 4.77 && \textbf{0.23} & \textbf{0.19} & \textbf{0.25}\\
				& GEM-CNN && 1.3 & 1.1 & 1.6 && 23.3 & 20.4 & 28.6 && \textbf{3.99} & \textbf{3.71} & \textbf{4.64} && 0.27 & 0.22 & 0.32\\
				\cmidrule{2-18}
				& SAGE-CNN$^\dagger$ && \gray{7.3} & \gray{6.9} & \gray{9.6} && \gray{117.4} & \gray{115.2} & \gray{151.3} && \gray{8.60} & \gray{8.29} & \gray{10.10} && \gray{1.45} & \gray{1.37} & \gray{1.91}\\
				& FeaSt-CNN$^\dagger$ && \gray{7.4} & \gray{7.4} & \gray{10.1} && \gray{119.6} & \gray{117.1} & \gray{161.1} && \gray{8.39} & \gray{8.33} & \gray{9.78} && \gray{1.48} & \gray{1.47} & \gray{2.01}\\
				& GEM-CNN$^\dagger$ && 1.3 & 1.1 & 1.6 && 23.3 & 20.9 & 28.5 && 4.00 & 3.67 & 4.65 && 0.26 & 0.22 & 0.32\\
				\bottomrule
			\end{tabular}
		}
	\end{center}
	\caption{Prediction accuracy on the idealised coronary arteries and left-main bifurcation models over the held-out test sets. We report normalised mean absolute error (NMAE), approximation error $\varepsilon$, as well as maximum and mean difference. CFD simulations act as ground truth. ($^\dagger$evaluated on randomly rotated test samples)}
	\label{tab:metrics}
\end{table}

\subsection{Equivariance}
Due to its signal representation, GEM-CNN is a globally $\SO{3}$-equivariant operator, i.e. the network output rotates with the input mesh orientation.
SAGE-CNN and FeaSt-CNN are not equivariant under $\SO{3}$ transformation.
This is because their signals are expressed in Euclidean space which makes the networks implicitly learn a reference orientation. To illustrate this, we compare the prediction accuracy on randomly rotated samples of the test sets (see Tab.~\ref{tab:metrics}). GEM-CNN's prediction accuracy is equivalent,
with only small numerical deviation due to discretisation of the activation functions.
In contrast, the performance of SAGE-CNN and FeaSt-CNN drops drastically, suggesting that they fail to predict useful WSS vectors on rotated samples.

\section{Discussion and conclusion}
We have demonstrated the feasibility of GCNs and mesh convolutional networks as surrogates for CFD blood flow simulation in arterial models. We found that GCNs effortlessly operate on the same meshes used in CFD simulations. Thus, shape parametrisation steps as proposed in other works \cite{SuZhang2020,GarleghiSamarasinghe2020,ItuRapaka2016}
may be omitted and GCNs can be seamlessly integrated into existing CFD workflows.
We noticed a significant speedup when using our method: inference in a new and unseen 3D geometry took less than 5 [s] including preprocessing, while CFD simulation took up to 15 [min].
The use of the same GCN architecture on two distinctly different datasets shows that our method does not require extensive task-specific fine-tuning and highlights its flexibility.

Our prediction results are based on two datasets which were synthesised with carefully chosen parameters to achieve a high level of realism. In future work, we aim to enrich our training and validation datasets with patient-specific artery geometries. This raises challenges regarding generalisation and uncertainty quantification, also with respect to the sensitivity of the proposed method to different meshing of the same lumen wall.
One limitation of our experiments is that we consider steady blood flow, whereas in real applications doctors might be interested in time-averaged WSS over a cardiac cycle of pulsatile blood flow. This may require the incorporation of patient-specific boundary conditions.
In the current experiments, boundary conditions were fixed across samples of both the single artery and bifurcating artery models and thus inherently encoded in the datasets. In future work, we will investigate to what extent the networks can be conditioned on user-supplied boundary conditions.

In conclusion, GCNs and mesh convolutional networks are a feasible approach to CFD surrogate modelling with potential applications in personalised hemodynamics estimation.

\section*{Acknowledgment}
This work is funded in part by the 4TU Precision Medicine programme supported by High Tech for a Sustainable Future, a framework commissioned by the four Universities of Technology of the Netherlands. Jelmer M. Wolterink was supported by the NWO domain Applied and Engineering Sciences VENI grant (18192).

\bibliographystyle{splncs04}
\bibliography{references}

\end{document}